\documentclass{article}
\usepackage{spconf,amsmath,graphicx}
\usepackage{multirow}

\title{GS-Net: Global Self-Attention Guided CNN for Multi-Stage Glaucoma Classification}
\name{Dipankar Das\ \ \ Deepak Ranjan Nayak\thanks{This work is supported by the Science and Engineering Research Board, Department of Science and Technology, Govt. of India under project No. SRG/2020/001460.}}
\address{Department of CSE, Malaviya National Institute of Technology, Jaipur, India}
\begin{document}
%
\maketitle
\begin{abstract}
Glaucoma is a common eye disease that leads to irreversible blindness unless timely detected. Hence, glaucoma detection at an early stage is of utmost importance for a better treatment plan and ultimately saving the vision. The recent literature has shown the prominence of CNN-based methods to detect glaucoma from retinal fundus images. However, such methods mainly focus on solving binary classification tasks and have not been thoroughly explored for the detection of different glaucoma stages, which is relatively challenging due to minute lesion size variations and high inter-class similarities. This paper proposes a global self-attention based network called GS-Net for efficient multi-stage glaucoma classification. We introduce a global self-attention module (GSAM) consisting of two parallel attention modules, a channel attention module (CAM) and a spatial attention module (SAM), to learn global feature dependencies across channel and spatial dimensions. The GSAM encourages extracting more discriminative and class-specific features from the fundus images. The experimental results on a publicly available dataset demonstrate that our GS-Net outperforms state-of-the-art methods. Also, the GSAM achieves competitive performance against popular attention modules.
\end{abstract}
\begin{keywords}
Multi-stage glaucoma classification, CNN, Global self-attention module, CAM, SAM.
\end{keywords}
\section{Introduction}
\label{sec:intro}
Glaucoma is a severe retinal disease that leads to optic nerve degeneration, resulting in irreversible and permanent vision loss. It is also known as the ``silent thief of sight" since many patients are unaware of their conditions in the early stages \cite{fu2018disc}. Glaucoma is caused by increased ocular pressure inside the eye, thereby causing damage to the optic nerve. Hence, the main clinical signs of glaucoma arise in the optic nerve region \cite{raghavendra2018deep}. Most glaucoma-related blindness can be avoided by early detection and treatment. Therefore, detecting glaucoma at an early stage is crucial to safeguard vision. \par
\begin{figure}[htp]
    \centering
    \includegraphics[width=8.5cm]{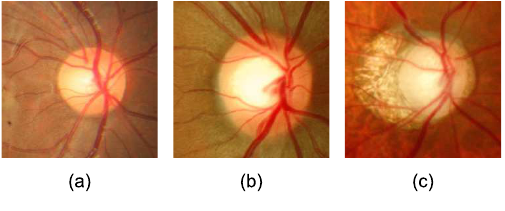}
    \caption{Sample retinal fundus images of different glaucoma stages (a) normal (b) early, and (c) advanced.}
    \label{fig:1}
\end{figure}
\begin{figure*}[h]
\centering
  \includegraphics[width=\textwidth]{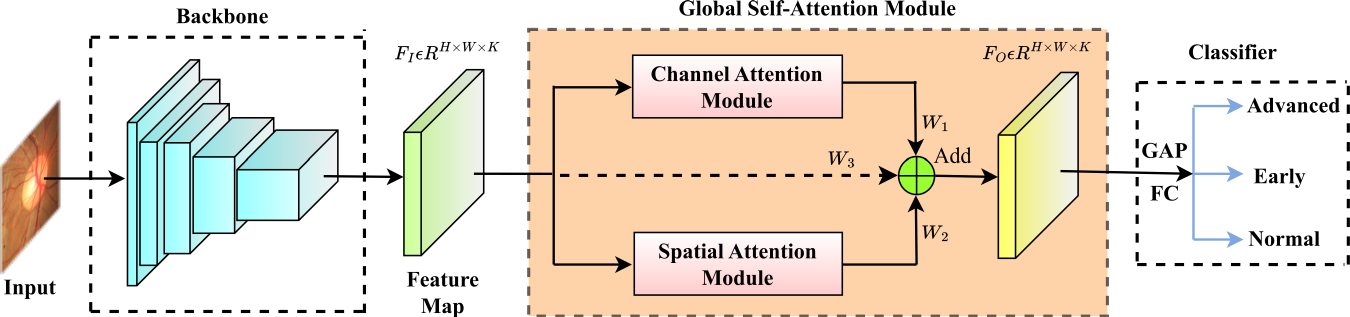}
  \caption{Structure of proposed global self-attention guided CNN (GS-Net).}
  \label{fig:2}
\end{figure*}
In practice, screening of glaucoma is performed using three tests: a function-based visual field test, an intraocular pressure (IOP) evaluation, and an optic nerve head (ONH) assessment test \cite{fu2018disc}. However, these tests are manual, which is tedious, time-consuming, and requires specialized observation. Therefore, it is essential to develop automated glaucoma screening methods to lessen the workload of ophthalmologists and deliver quick and accurate decisions.
In the literature, several methods~\cite{fu2018joint,cheng2013superpixel} have been reported based on the segmentation of the optic disc and cup from fundus images followed by the measurement of clinical values to detect glaucoma. However, these methods depend on segmentation accuracy and suffer from the problem of low sensitivity. Later, many machine learning-based methods \cite{maheshwari2016automated,maheshwari2019automated} have been proposed using different hand-crafted features such as wavelet features, non-linear entropy features, local binary patterns, etc., and standard classifiers. Although these methods have shown good results, they require choosing suitable feature extraction and classification methods. To avoid these problems, researchers have shown significant interest in designing CNN-based models \cite{chen2015glaucoma, li2016integrating, pal2018g, tian2022gc}, which have been effective in learning high-level feature representations from fundus images. However, the aforementioned methods are focused towards solving binary classification tasks and can not decide the different stages of glaucoma. Multi-stage glaucoma classification is crucial and relatively challenging due to high inter-class similarities and minute variations in lesion size. A few typical samples of fundus images of different glaucoma stages are depicted in Fig.~\ref{fig:1}. Limited methods have been proposed till date for the classification of different stages of glaucoma. For instance, in \cite{parashar20212} and \cite{parashar2020automatic}, authors used machine learning-based approaches to classify the various stages of glaucoma. Recently, in \cite{ahn2018deep}, a CNN-based method was proposed for different glaucoma stage classification. However, these conventional CNNs face difficulty in learning fine-grained features from the lesion regions. 

Attention mechanisms have emerged as a key component in modern CNN architectures, allowing such architectures to emphasize on salient features and simultaneously suppress irrelevant information \cite{hu2018squeeze, woo2018cbam, tian2022gc}. Among them, self-attention has gained remarkable attention due to its ability to learn global feature dependencies \cite{zhang2019self}. Inspired by this, in this study, we present a global self-attention module (GSAM) to capture more salient and discriminative features from fundus images. The GSAM is integrated with a backbone model and is referred to as GS-Net. 
The major contributions are summed up as follows: $(1)$ We present a novel attention-based network GS-Net for multi-stage glaucoma classification, $(2)$ We introduce GSAM after a backbone which leverage the model's capability in capturing feature inter-dependencies across both channel and spatial dimensions,
$(3)$ We compare the effectiveness of our GSAM against cutting-edge attention modules (e.g., SE\cite{hu2018squeeze}, CBAM \cite{woo2018cbam}, GAB \cite{he2020cabnet} and GC \cite{cao2019gcnet}) as well as state-of-the-art glaucoma detection schemes, and $(4)$ We conduct ablation studies to analyze the effect of each component of GSAM. 
\section{Proposed Methodology}
\label{sec:format}
The detailed architecture of our GS-Net is illustrated in Fig.~\ref{fig:2}, which contains three major parts: a backbone network, a global self-attention module, and a classifier. 
\subsection{Backbone}
As depicted in Fig.~\ref{fig:2}, GS-Net takes a fundus image as input and then, it uses a backbone network to derive the high-level feature maps $F_{I}\in R^{H\times W\times K}$, where $K$, $W$, and $H$ denote the number of channels, width, and height, respectively. The backbone network is a CNN model pre-trained using ImageNet dataset. $F_{I}$ is acquired from the last convolutional layer of the backbone model and is utilized as the input to the GSAM. 
\subsection{Proposed Global Self-Attention Module (GSAM)}
\begin{figure}[!htb]
	\centering
	\includegraphics[width=\columnwidth,height=6cm]{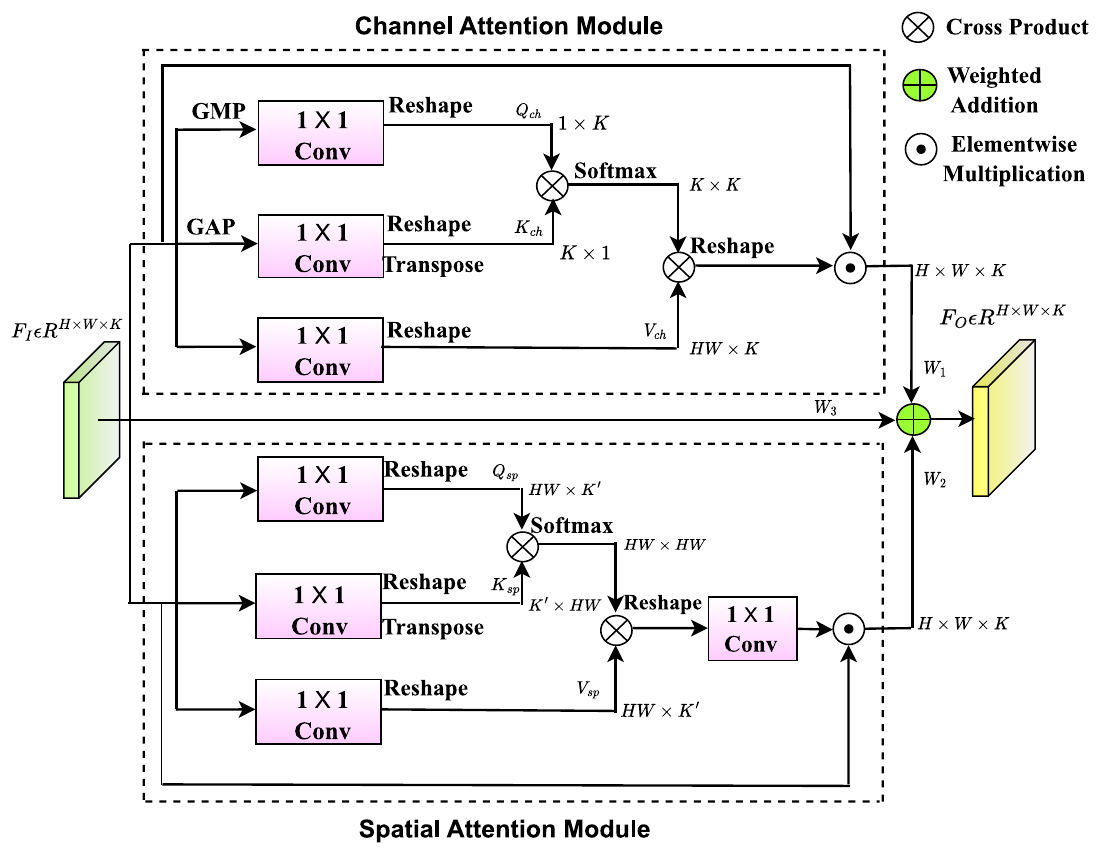}
	\caption{Illustration of proposed GSAM.}
	\label{fig:3}
\end{figure}

The GSAM is proposed to learn more salient and class-specific discriminative features from the lesion areas. Unlike the self-attention methods reported in \cite{zhang2019self} and \cite{wang2018non} which enables interaction among the spatial positions only, our GSAM introduces two parallel self-attention modules, channel attention module (CAM) and spatial attention module (SAM) to enable interaction among different channels and spatial positions. Finally, the attention feature maps obtained from each module are fused with the input feature map using a trainable fusion strategy to acquire a global attention feature map. Fig.~\ref{fig:3} illustrates the overview of GSAM.
\subsubsection{Channel Attention Module (CAM) }
The goal of channel attention is to determine feature interdependencies across the channel dimension and hence, we introduce a CAM to evaluate the significance of each channel. 
The input feature map $F_{I}\in R^{H\times W\times K}$ is fed to three branches of CAM simultaneously. In the first branch, a global max pooling (GMP) is performed, followed by a $1 \times 1$ Conv block, and reshape operation (RO) to obtain a query tensor $Q_{ch}\in R^{1\times K}$. While in the second branch, global average pooling (GAP) and a $1 \times 1$ Conv block are introduced followed by RO and transpose operation to obtain the key tensor $K_{ch}\in R^{K\times 1}$. Then, the $Q_{ch}$ and $K_{ch}$ are multiplied followed by softmax activation to obtain the attention weights $F_{ch}'\in R^{K\times K}$, which is defined in (\ref{eq:1}).
\begin{equation}\label{eq:1}
F_{ch}'= \sigma (K_{ch} \otimes Q_{ch})
\end{equation}
where, $\sigma$ and $\otimes$ denote softmax and cross-product, respectively. In the third branch, a $1 \times 1$ Conv block is applied followed by a reshape operation to produce the value tensor $V_{ch}\in R^{HW\times K}$. Next, $V_{ch}$ and $F_{ch}'$ are multiplied followed by RO to obtain the channel-wise attention feature tensor $F_{ch}''\in R^{H \times W\times K}$, which is expressed in (\ref{eq:2}).
\begin{equation}\label{eq:2}
F_{ch}''= V_{ch} \otimes F_{ch}'
\end{equation}
At last, the feature map $F_{ch}''$ is multiplied element-wise with $F_{I}$ to yield the global channel-wise attention map $F_{ch}$ as
\begin{equation}
F_{ch}= F_{ch}'' \odot F_{I}
\end{equation}
where, $\odot$ denotes element-wise multiplication operation. It is worth noting that each $1 \times 1$ Conv block includes a layer normalization and swish activation layer.
\subsubsection{Spatial Attention Module (SAM)}
The optic disc and cup regions in the fundus images of glaucoma patients highly vary in size. Hence, we introduce SAM to establish the relationship among each position within the feature map, which is related to self-attention \cite{zhang2019self} method. 
The SAM takes $F_{I}\in R^{H\times W\times K}$ as input to its three branches. In the first and last branch, a $1 \times 1$ Conv block with $K'$ channels followed by RO is implemented to obtain the query tensor $Q_{sp}\in R^{HW\times K'}$ and value tensor $V_{sp}\in R^{HW\times K'}$, respectively, where $K' = K/2$. 
While the second branch includes a $1 \times 1$ Conv block with $K'$ channels followed by RO and transpose operation to obtain the key tensor $K_{sp}\in R^{K'\times HW}$. Then, the multiplication between $Q_{sp}$ and $K_{sp}$ is performed followed by softmax activation to yield the spatial attention weights $F_{sp}'\in R^{HW\times HW}$ as
\begin{equation}
    F_{sp}'= \sigma (Q_{sp} \otimes K_{sp})
\end{equation}
Next, the $F_{sp}'$ is multiplied with the value tensor $V_{sp}$ which is followed by RO and a $1 \times 1 $ Conv block with $K$ channels to obtain the spatial-wise attention feature tensor $F_{sp}''\in R^{H \times W\times K}$. 
\begin{equation}
F_{sp}''= Conv(F_{sp}' \otimes V_{sp})
\end{equation}
Finally, the global spatial-wise attention feature map $F_{sp}$ is obtained by performing element-wise multiplication between $F_{sp}''$ and $F_{I}$, which is expressed in (\ref{eq:6}).
\begin{equation}\label{eq:6}
F_{sp}= F_{sp}'' \odot F_{I}
\end{equation}

\subsection{Fusion Strategy}
The output attention feature maps from each module are fused with the original feature maps $F_{I}$  using their weighted sum with weights $W_1$, $W_2$, and $W_3$, to obtain the global attention feature map $F_{O}\in R^{H\times W\times K}$. This allows the network to emphasize on salient lesion regions while ignoring irrelevant regions, thereby, capturing more class-specific features. Mathematically, it can be formulated as
\begin{equation}
    F_{O}= W_{1}F_{ch} \oplus W_{2}F_{sp} \oplus W_{3}F_{I} 
\end{equation}
where, $W_{1}$, $W_{2}$, and $W_{3}$ denote the learnable scalar weights.
 
\subsection{Classifier}
We employ GAP followed by a fully-connected (FC) layer of three neurons to detect different stages of glaucoma such as normal, early and advanced.
\
\section{Experiments and results}
\label{sec:pagestyle}
\subsection{Dataset}
To validate our proposed method as well as existing approaches, we use a publicly available Harvard Dataverse V1 dataset \cite{ahn2018deep}, which contains a total of 1524 fundus images from three classes:  normal (786 images), early-stage glaucoma (289 images), and advanced stage glaucoma (467 images). These fundus images are captured through the non-mydriatic auto fundus camera AFC 330 at Kim's Eye Hospital, South Korea.
\subsection{Experimental Setup}
To derive a fair comparison with state-of-the-art methods, we adopt a similar data split strategy as reported in \cite{ahn2018deep}, i.e., we split the dataset into train, validation, and test sets with 754, 324, and 464 images, respectively. The input fundus images are resized to $224\times224$ pixels. Further, we use several data augmentation techniques, namely, random rotation, scaling, horizontal flip, and vertical flip to avoid overfitting. We train each model for 50 epochs with an initial learning rate of 0.005 and a batch size of 16. The categorical cross-entropy loss and Adam optimizer are utilized to train all models. The proposed model is designed and implemented using the Keras library and TensorFlow as the back end. To evaluate all models, we use different performance measures like accuracy (Acc), F1-score (F1), and AUC.
\subsection{Results}\label{re}
To analyze the efficacy of backbones in our proposed model, we evaluate the performance of various popular ImageNet pre-trained CNN architectures such as ResNet-50 \cite{he2016deep}, InceptionV3 \cite{szegedy2016rethinking}, MobileNet \cite{howard2017mobilenets}, EfficientNetB1 \cite{tan2019efficientnet}, and DenseNet-121 \cite{huang2017densely} and the results are given in Table~\ref{table:1}. 
\begin{table}[h!]
\begin{center}
\caption{Comparison of different backbones and popular attention modules on Harvard Dataverse V1 dataset}
\resizebox{\columnwidth}{!}
{\begin{tabular}{llccc} 
 \hline
Backbone & Attention & Acc (\%) & F1 (\%) & AUC \\ 
 \hline \hline 
  \multirow{6}{*}{\begin{tabular}[c]{@{}l@{}}ResNet-50 \end{tabular}} 
 &None&81.25&80.98&0.9259\\
 &SE~\cite{hu2018squeeze} &81.89&81.50&0.9217\\
 &GC~\cite{cao2019gcnet} &81.89&81.50&0.9217\\
 &GAB~\cite{he2020cabnet} &82.75&82.46&0.9259\\
 &CBAM~\cite{woo2018cbam} &82.75&82.46&0.9259\\
 &Ours &\textbf{83.40}&\textbf{83.27}&\textbf{0.9336}\\
 \hline
  \multirow{6}{*}{\begin{tabular}[c]{@{}l@{}}InceptionV3 \end{tabular}} 
&None&80.60&80.42&0.9244\\
 &SE~\cite{hu2018squeeze}&80.60&80.42&0.9244\\
 &GC~\cite{cao2019gcnet}&81.03&80.99&0.9222\\
 &GAB~\cite{he2020cabnet} &81.25&80.98&0.9259\\
 &CBAM~\cite{woo2018cbam} &81.03&80.99&0.9222\\
 &Ours &\textbf{81.46}&\textbf{81.40}&\textbf{0.9279}\\
 \hline
 \multirow{6}{*}{\begin{tabular}[c]{@{}l@{}}MobileNet \end{tabular}} 
 &None&78.66&76.90&0.9216\\
 &SE~\cite{hu2018squeeze}&79.74&79.06&0.9066\\
 &GC~\cite{cao2019gcnet} &78.66&76.90&0.9216\\
 &GAB~\cite{he2020cabnet}&80.17&79.43&0.9238\\
 &CBAM~\cite{woo2018cbam} &79.74&79.06&0.9066\\
 &Ours &\textbf{80.60}&\textbf{80.42}&\textbf{0.9244}\\
 \hline
   \multirow{6}{*}{\begin{tabular}[c]{@{}l@{}}DenseNet-121 \end{tabular}}
 &None&83.18&83.13&0.9357\\
 &SE~\cite{hu2018squeeze}&83.18&83.13&0.9357\\
 &GC~\cite{cao2019gcnet} &83.40&83.27&0.9336\\
 &GAB~\cite{he2020cabnet} &84.48&84.34&0.9114\\
 &CBAM~\cite{woo2018cbam} &83.40&83.27&0.9336\\
 &Ours &\textbf{84.91}&\textbf{84.55}&\textbf{0.9454}\\
 \hline
  \multirow{6}{*}{\begin{tabular}[c]{@{}l@{}}EfficientNetB1 \end{tabular}} 
  &None&79.74&79.21&0.9056\\
 &SE~\cite{hu2018squeeze}&80.17&79.43&0.9238\\
 &GC~\cite{cao2019gcnet} &80.17&79.43&0.9238\\
 &GAB~\cite{he2020cabnet} &80.60&80.42&0.9244\\
 &CBAM~\cite{woo2018cbam} &80.17&79.43&0.9238\\
 &Ours &\textbf{80.60}&\textbf{80.42}&\textbf{0.9244}\\
 \hline

\end{tabular}}
\label{table:1}
\end{center}
\end{table}
\begin{table}[h!]
\begin{center}
\caption{Results of different components of our GSAM}
\begin{tabular}{lccc} 
 \hline
Method & Acc (\%) & F1 (\%) & AUC \\ 
 \hline \hline 
 Baseline &83.18&83.13&0.9357\\
 Baseline + CAM&84.69&84.35&0.9443\\
 Baseline + SAM &84.48&84.12&0.9415\\
 Ours &\textbf{84.91}&\textbf{84.55} &\textbf{0.9454}\\
 \hline
 \end{tabular}
\label{table:2}
\end{center}
\end{table}
\begin{table}[h!]
\begin{center}
\caption{Performance comparison with state-of-the-art CNN-based glaucoma detection methods on Harvard Dataverse V1 dataset}
\resizebox{\columnwidth}{!}{\begin{tabular}{ lccc } 
 \hline
 Method& Acc (\%) & F1 (\%) & AUC \\
 \hline \hline 
 4-Layer CNN \cite{chen2015glaucoma} & 75.64&75.84  &0.8900 
 \\
 Customized CNN \cite{ahn2018deep}& 78.45&78.42  &0.9049 
 \\
InceptionV3 \cite{ahn2018deep} &80.60&80.42&0.9244
\\
 Customized CNN \cite{raghavendra2018deep} &78.01  & 79.02 &0.9097 
 \\
 ResNet-50 + GAB+ CAB \cite{tian2022gc}& 82.75&82.46&0.9259
 \\
 Ours (GS-Net) &\textbf{84.91}&\textbf{84.55}&\textbf{0.9454}
 \\
 \hline
 \end{tabular}}
 \label{table:3}
\end{center}
\end{table}
It can be observed that DenseNet121 outperforms others with a higher classification accuracy of 83.18\% and is therefore used as the backbone in our GS-Net model. Moreover, to test the effectiveness of the GSAM, we compared it with popular attention modules such as SE~\cite{hu2018squeeze}, CBAM~\cite{woo2018cbam}, GAB~\cite{he2020cabnet} and GC~\cite{cao2019gcnet} by adopting different backbone networks as shown in Table~\ref{table:1}. It can be observed that our proposed GSAM performs better than other attention modules with all the backbones and the GSAM with DenseNet-121 achieves the highest classification accuracy of 84.91\%. It is noteworthy that all these experiments are evaluated under the same experimental setup.

 Ablation studies are further carried out to explore the impact of each component of the proposed GSAM. Table~\ref{table:2} demonstrates that the model with CAM and SAM in standalone obtained a relatively lower performance, while the combination of both improves the classification performance. For this experiment, we adopt the best-performing backbone i.e., DenseNet-121.  Further, we validate the existing CNN-based glaucoma detection methods on the Harvard Dataverse V1 dataset under a similar training environment and compare their performance with our GS-Net model as shown in Table~\ref{table:3}. It is evident that GS-Net attains superior classification performance as compared to other state-of-the-art approaches. Further, it can be noticed that our method outperforms the attention-based approach ResNet-50+GAB+CAB~\cite{tian2022gc}. The superior performance of our GS-Net is majorly due to its ability to learn fine-grained and discriminative lesion features using GSAM. 

\section{Conclusion}
\label{sec:typestyle}
We propose a novel global self-attention guided network called GS-Net for multi-stage glaucoma classification using retinal fundus images. The proposed attention module GSAM is coupled with a backbone and helps to explore global feature dependencies. Hence, it facilitates extracting more discriminative and class-specific features from the lesion regions, improving classification performance. The proposed GS-Net can be built and learned end-to-end, showing its suitability for real-time practical application. The experimental results on an openly available dataset indicate the superiority of our method compared to state-of-the-art glaucoma detection methods. Furthermore, the comparison with popular attention modules demonstrates the efficacy of our proposed GSAM. 



\end{document}